\documentclass[letterpaper, 10pt, journal, twoside]{IEEEtran}
\IEEEoverridecommandlockouts
\usepackage{graphicx} % for pdf, bitmapped graphics files
\usepackage{subfigure}
\usepackage{amsmath} % assumes amsmath package installed
\usepackage{amssymb}  % assumes amsmath package installed
\usepackage{booktabs}
\usepackage{caption}
\usepackage{multirow}

\usepackage{hyperref}

\usepackage[normalem]{ulem}
\useunder{\uline}{\ul}{}

\title{RO-MAP: Real-Time Multi-Object Mapping with Neural Radiance Fields}

\author{Xiao Han, Houxuan Liu, Yunchao Ding and Lu Yang %
	\thanks{Manuscript received: April 12, 2023; Revised June 27, 2023; Accepted July 24, 2023. This paper was recommended for publication by Editor Markus Vincze upon evaluation of the Associate Editor and Reviewers' comments. This work was supported by NSFC (No. 61871074).}% Use only for final RAL version
	\thanks{All authors are with the School of Automation Engineering, University of Electronic Science and Technology of China, Chengdu, China. Lu Yang is the corresponding author {\tt\footnotesize yanglu@uestc.edu.cn.}}%
	\thanks{Digital Object Identifier (DOI): see top of this page.}%
}
\begin{document}
	
\maketitle

\begin{abstract}
Accurate perception of objects in the environment is important for improving the scene understanding capability of SLAM systems. In robotic and augmented reality applications, object maps with semantic and metric information show attractive advantages. In this paper, we present RO-MAP, a novel multi-object mapping pipeline that does not rely on 3D priors. Given only monocular input, we use neural radiance fields to represent objects and couple them with a lightweight object SLAM based on multi-view geometry, to simultaneously localize objects and implicitly learn their dense geometry. We create separate implicit models for each detected object and train them dynamically and in parallel as new observations are added. Experiments on synthetic and real-world datasets demonstrate that our method can generate semantic object map with shape reconstruction, and be competitive with offline methods while achieving real-time performance (25Hz). The code and dataset will be available at: \href{https://github.com/XiaoHan-Git/RO-MAP}{https://github.com/XiaoHan-Git/RO-MAP}

\end{abstract}


\section{INTRODUCTION}

\IEEEPARstart{V}{ision}-based Simultaneous Localisation and Mapping (SLAM) is an important research problem in the field of robotics, and has achieved remarkable advances in the past decade. Previous studies \cite{orb,engel2014lsd,pumarola2017pl} concentrated on providing accurate ego-motion estimation and reconstructing scene maps. However, the sparse or dense maps constructed by these methods only contain metric information, which limits their application in complex tasks \cite{martins2020extending,wu2021object} that require scene understanding. The development of deep learning has paved the way for introducing semantic information into SLAM, and object SLAM that incorporates detection \cite{redmon2016you} or semantic segmentation \cite{he2017mask} has attracted the interest of many researchers.

Different from pure geometric maps, object SLAM utilizes additional semantic observations to localize and reconstruct objects in the scene, and the generated object maps can serve downstream tasks.
However, a crucial issue is how to effectively represent objects. Some research that use only RGB cameras have explored simple geometric primitives, such as cuboids \cite{yang2019cubeslam,wu2020eao}, ellipsoids \cite{nicholson2018quadricslam,TianAccurate,liao2022so}, and superquadrics \cite{han2022sq}. 
These compact representations contain fundamental information of objects such as category, size, and pose.
They can serve as semantic landmarks for localization and navigation \cite{martins2020extending}, and have shown advantages in relocalization \cite{qian2022towards} and long-term operation of SLAM systems.
However, these geometric primitives do not capture shape and texture information of objects, which poses a challenge for monocular-based methods.

\begin{figure}[t]
	\centering
	\includegraphics[width=\linewidth]{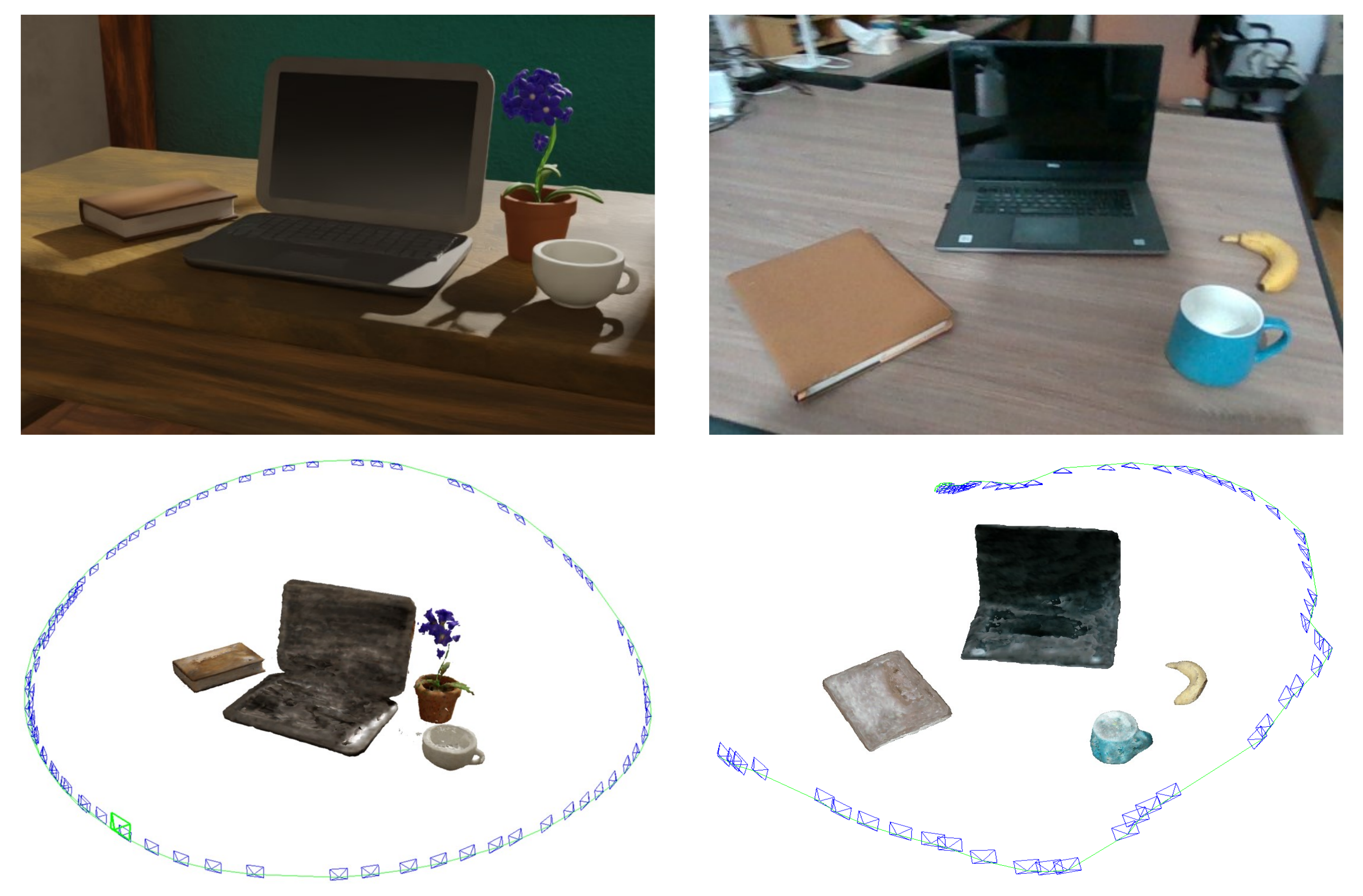}
	\caption{RO-MAP can localize and reconstruct objects in the scene online. Each object instance is represented by a NeRF, which implicitly learns dense geometry from monocular input.}
	\label{fig1}
\end{figure} 

Object shape reconstruction is another widely studied problem. Some studies have explored various dense object representations using additional depth sensors, such as surfels \cite{runz2018maskfusion} and signed distance function (SDF) \cite{xu2019mid,mccormac2018fusion}.
Furthermore, it is popular to use learnable compact shape embeddings to represent objects.
Recent works \cite{sucar2020nodeslam,shan2021ellipsdf,wang2021dsp,zou2022objectfusion} have used neural networks to learn category-level shape priors and optimized the object shape codes in latent space by matching image or depth observations, which are then decoded into voxel grids \cite{sucar2020nodeslam} or implicit functions \cite{shan2021ellipsdf,wang2021dsp,zou2022objectfusion}. These methods can generate dense and complete object reconstructions from partial observations, but they are limited by the categories of pre-learned priors and cannot handle arbitrary geometric shapes.
A natural question is whether we can reconstruct objects using only a monocular camera and without any geometric priors.
%Therefore, we pose the question: can we reconstruct objects using only a monocular camera and without any geometric priors?
Neural Radiance Fields (NeRF) \cite{mildenhall2021nerf} are suitable object representations. With the help of volume rendering and the powerful fitting ability of MLP, NeRF can implicitly learn 3D geometry from RGB images.
Recently, its successful applications\cite{sucar2021imap,zhu2022nice} in SLAM have demonstrated its strong potential.

In this work, we propose an online pipeline for reconstructing multiple objects from monocular videos, which consists of two loosely coupled components.
The first component is a lightweight object SLAM built upon the ORB-SLAM2 \cite{orb} framework.
We use instance segmentation to detect objects in the scene and estimate their size and pose, and a robust data association algorithm ensures that multi-view observations are correctly associated to objects.
The second component is a multi-object reconstruction system, where each object instance is represented by a NeRF and receives new observations in real-time for incremental training. We propose an efficient loss function tailored for objects to speed up convergence and reduce depth ambiguity caused by RGB-only images. Furthermore, our CUDA implementation based on the tcnn framework \cite{tiny-cuda-nn} ensures real-time performance. The average training time per object is about 2 seconds on a single GPU. Comprehensive experiments on synthetic and collected real-world datasets demonstrate the effectiveness of our method.

The contributions of this work are as follows:
\begin{itemize}
	
	\item We present, to the best of our knowledge, the first 3D prior-free monocular multi-object mapping pipeline that can localize and reconstruct objects in the scene.
	
	\item We propose an efficient loss function for objects, combined with a high performance CUDA implementation, enabling the system to have real-time performance.
	
	\item We evaluate the effectiveness of the proposed method on both synthetic and real-world datasets. In addition, the code and datasets are available.
\end{itemize}

\section{RELATED WORK}

\subsection{Object SLAM}

The earliest object SLAM can be traced back to 2013 when Salas et al. \cite{salas2013slam} first treated objects in the scene as landmarks. They matched depth observations extracted by pre-trained detectors with known object models, and continuously refined camera poses and object map through pose-graph optimization. However, the requirement of having geometric models of all object instances beforehand limits the applicability of this approach. Subsequent works along the RGB-D direction turned to online reconstruction from scratch.
Fusion++ \cite{mccormac2018fusion} leverages a 2D instance segmentation network to extract depth observations belonging to objects, and fuses them across multiple views to generate TSDF reconstructions. MaskFusion \cite{runz2018maskfusion} further extend to dynamic scenes.
Some studies lie between using object instance models and reconstructing geometric shapes from scratch, exploring learning-based object shape priors. Sucar et al. \cite{sucar2020nodeslam} used CAD models of objects from the same category to train a variational encoder. The latent code used to represent shape can generate complete object reconstruction with only partial depth observation. Similarly, ObjectFusion \cite{zou2022objectfusion} further improves the generalization of deep shape embeddings, capable of adapting to multiple object categories using only one encoder-decoder network.

Unlike the above studies that focus on dense object reconstruction, researchers have also shown interest in constructing object maps using simple geometric primitives that only contain pose and size information. Yang et al. \cite{yang2019cubeslam} proposed CubeSLAM, which infers 3D cuboids of objects from multi-view observations using 2D detection boxes. Compared to learning-based methods, CubeSLAM has much less computational cost. In addition, quadrics have also been used to represent objects due to their compact perspective projection model. Nicholson et al. \cite{nicholson2018quadricslam} first introduced them into object SLAM, and optimized their parameters by minimizing reprojection errors in different viewpoints. 
In this work, we decouple the estimation of object shape and pose. Our lightweight object SLAM utilizes cuboids to represent objects, and estimates their pose and size using multi-view observations and sparse point clouds.

\begin{figure*}[ht]
	\centering
	\includegraphics[width=\linewidth]{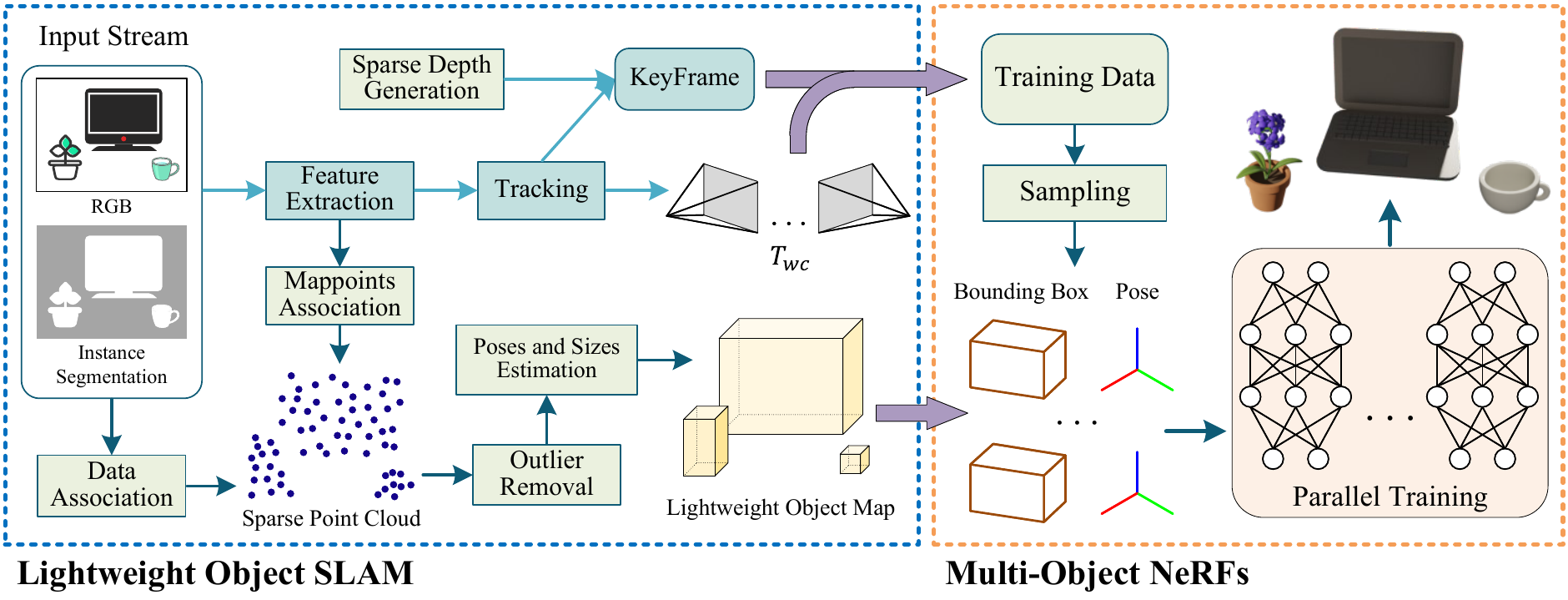}
	\caption{Pipeline overview. RO-MAP consists of a lightweight object SLAM and a multi-object NeRF system. They correspond to the bounding box estimation (pose and size) and shape reconstruction of objects, respectively.}
	\label{fig2}
\end{figure*} 

\subsection{NeRFs and NeRF-based SLAM}

Neural radiance fields have emerged as popular implicit representations in computer vision. They use a multi-layer perceptron (MLP) to represent a scene and leverage volume rendering to implicitly learn the geometry and appearance information from multi-view images. Since its introduction, NeRF has been widely used in various computer vision tasks \cite{mildenhall2021nerf,kundu2022panoptic}.
These methods demonstrate remarkable performance but require a large amount of training time.
Recently, Muller et al. \cite{muller2022instant} proposed a multi-resolution hash encoding, which significantly reduces the training time of NeRF, making its online application possible.

Due to the advantage of NeRF to implicitly represent 3D scenes, several works have introduced it into SLAM. 
iMAP \cite{sucar2021imap} is the first complete SLAM system based on NeRF, which utilizes a compact MLP-based map representation and performs simultaneous tracking and mapping in two parallel threads. NICE-SLAM \cite{zhu2022nice} replaces a single MLP with a hierarchical feature grid, combines with a pre-trained decoder to achieve larger scenes and faster convergence. Compared to iMAP, it updates only the visible grid features at each step, effectively solving the forgetting problem. Subsequent works \cite{rosinol2022nerf,zhu2023nicer} have made further improvements, including the integration with traditional voxel grids \cite{yang2022vox} and different shape representations \cite{ming2022idf}. In contrast to these methods that focus on dense reconstruction of the scene, our approach emphasizes object instances with semantic meaning. Each object is represented by a separate NeRF model, which are trained online and in parallel to generate dense object map. The work most similar to ours is \cite{abou2022implicit}, but it focuses on analyzing the effect of observation quality on reconstruction results. In addition, vMAP \cite{kong2023vmap} also models objects separately while reconstructing the scene, but it uses RGB-D observations.

\section{System Overview}

Fig. \ref{fig2} shows an overview of the proposed method. The pipeline consists of two main components, a lightweight object SLAM and a multi-object NeRF system. Given a monocular input stream that includes RGB images and instance segmentations, our object SLAM system simultaneously estimates camera frame poses and localizes objects in the scene. We leverage both semantic information from instance segmentation and geometric information from sparse point clouds associated with objects to perform data association and object pose and size estimation. The results and original image inputs are fed into the multi-object reconstruction system. In this part, each object instance is represented by a separate NeRF model. They receive new observations in real-time and are trained in parallel. We use the marching cubes algorithm \cite{lorensen1987marching} to extract visual 3D meshes and transform them to the global coordinate system through object poses, thus constructing a complete dense object map.

\section{Lightweight Object SLAM}

Our object SLAM is implemented based on ORB-SLAM2 \cite{orb}. For monocular input, the system initially extracts ORB features and performs inter-frame matching. As we only consider static scenes, the camera pose estimation is consistent with the original ORB-SLAM2, i.e. only the traditional reprojection error is used. Simultaneously, matched image features are triangulated to generate sparse point clouds. We associate the sparse point clouds with object instance segmentation, effectively utilizing them in subsequent object data association and pose estimation.
The object association strategy comprises two parts: consecutive association and non-consecutive association. The former calculates the intersection over union (IoU) of object 2D bounding boxes obtained from instance segmentation in consecutive frames. The latter utilizes a parameter statistical test based on object's sparse point clouds to handle isolated object observations in time series and merge duplicate object landmarks. Since data association is not the focus of this paper, please refer to our previous work \cite{han2022sq} for more details. After associating the latest object observations with landmarks, we employ a lightweight, hand-crafted method for object localization instead of learning-based methods.

\subsection{Outlier Removal}
After extracting image features, the feature points that are covered by the instance mask will be associated to objects. The sparse point cloud generated by triangulating these points during the tracking process is used for roughly representing the position of objects. However, due to measurement noise and occlusion, the associated sparse point cloud often contains many outliers that do not belong to objects. We employ the Extended Isolation Forest (EIF) \cite{hariri2019extended} to remove outliers and maintain a sparse point cloud that accurately fits objects. Specifically, EIF recursively partitions the sample space using a plane with a random slope, gradually reducing the number of samples in each enclosed space until each sample is isolated or the depth limit is reached. Obviously, the points that are located on the object surface after multi-view observations tend to be dense, and require more steps to be isolated. We remove those points that are isolated after very few steps, which are likely to be outliers.

\subsection{Pose and Size Estimation}

We represent objects using cuboids and assume that objects are always stationary and placed on a support, with roll and pitch angles fixed at zero, so that only the translation $\mathbf{t}$ and yaw angle $\theta$ need to be estimated. First, we directly compute the center of the filtered point cloud $P^W$ to estimate the translation $\mathbf{t}$ as follows:
\begin{equation}
\mathbf{t}  = \frac{\max\left ( P^W \right )   + \min\left ( P^W \right ) }{2} 
\end{equation}

For object rotation, the simple and effective Principal Component Analysis (PCA) method is considered first. We project the 3D sparse point cloud onto the horizontal plane and then calculate its dominant orientation using PCA as the corresponding rotation matrix. However, this method performs poorly for cuboid-shaped objects such as books and keyboards, as the extracted main direction deviates significantly from the ideal orthogonal edges. This results in inaccurate object pose estimation and further affects the subsequent shape reconstruction. We combine a line feature alignment method based on object appearance to improve the robustness of rotation estimation. 
Specifically, we first project the three orthogonal edges $l_{i}(i\in1,2,3)$ of the object bounding box onto the image, then extract line features \cite{akinlar2011edlines} and select those that have similar slopes to the projected line segments as observations. The accumulative angle error between the extracted line segments $l_{di}$ and the projected line segments $l_{oi}$ is optimized to estimate yaw $\theta$. The optimization function is defined as follows:
\begin{gather}
\theta ^* = \mathop{\arg\min}_{\theta}\sum_{i=1}^{3} \left \| g(l_{oi})- g(l_{di})\right \| ^2 \\
l_{oi} = KT_{wc}^{-1} \left (  R\left ( \theta\right ) l_i + \mathbf{t}  \right ), \  i\in \left \{ 1,2,3 \right \} 
\end{gather}
where $g\left ( \cdot \right ) $ calculates the slope of line segment. $T_{wc}$ represent the camera pose and $K$ is the camera intrinsic matrix. A good initial value is crucial for this nonlinear optimization problem. We uniformly sample from $-45^\circ $ to $45^\circ $ with an interval of 5 degrees, and select the sample with the minimum error as the initial value for optimization. Finally, we obtain the sparse point cloud $P^O$ transformed to the object coordinate system by the estimated object pose, and directly calculate the size $\mathbf{a}  = \left [ a_x,a_y,a_z  \right ] ^T$ as follows:
\begin{equation}
\mathbf{a}  = \frac{\max\left ( P^O \right )   - \min\left ( P^O \right ) }{2} 
\end{equation}

\section{Multi-Object Reconstruction System}

After estimating bounding boxes and camera poses in object SLAM, we use NeRF to implicitly learn the dense geometry of objects.
When a new object instance is detected, we initialize a new NeRF model, which consists of a multi-resolution hash encoding \cite{muller2022instant} and a single-layer MLP. Unlike some methods that reconstruct the whole scene with NeRFs, our model only needs to represent a single object, which allows us to use tiny network structures and accelerate training speed considerably. Moreover, we leverage multi-threading to train models in parallel, thus further improving system efficiency .

\subsection{Training}

\subsubsection{Data}

\begin{figure}[t]
	\centering
	\includegraphics[width=0.95\linewidth]{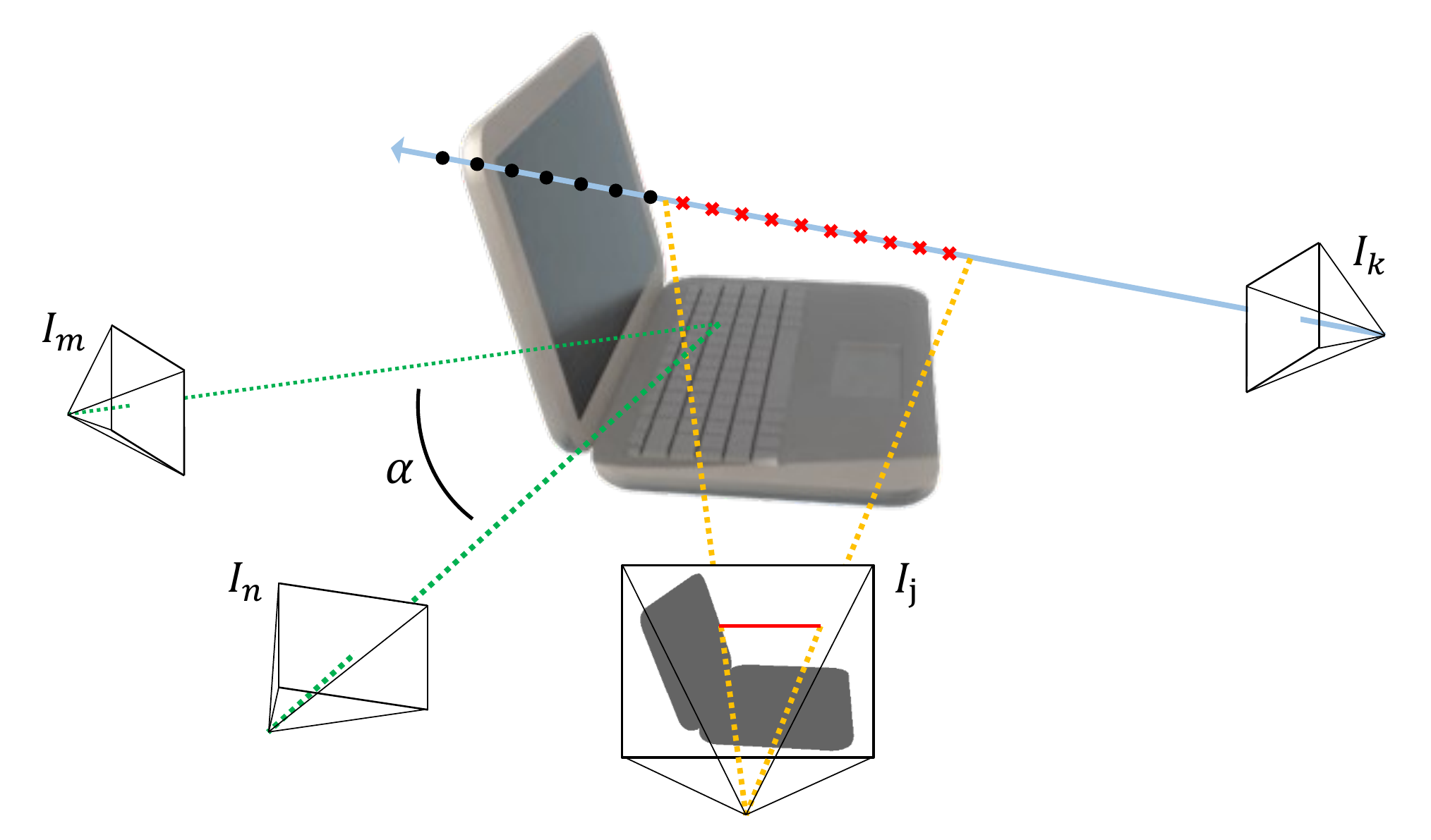}
	\caption{(a) The training data will be updated if the rotation angle between the current image $I_n$ and the last updated image $I_m$ is larger than $\alpha$. (b) Thanks to the zero density learned by background rays in image $I_j$, the rays pointing to the object in image $I_k$ can quickly focus on the object surface.}
	\label{fig3}
\end{figure}

Since there are small viewpoint changes between adjacent frames in SLAM, using all images for training would introduce a lot of redundant information. We only use those images that are chosen as keyframes in the tracking process. 
Besides the original RGB images and instance masks, we also reproject the sparse point clouds associated with objects onto the images, and the resulting sparse depth maps can serve as additional supervision during training.
This facilitates the model to learn accurate geometry.
For all object instances, they have different training data and different appearance times. We implement an incremental update method for the training data to handle each model separately. As shown in Fig. \ref{fig3}, assuming that the last updated image is $I_m$ and the currently observed object image is $I_n$, we calculate their relative rotation angles with respect to the object as follows:
\begin{equation}
\alpha = \arccos \left (   \frac{(\mathbf{t}_{I_m}-\mathbf{t}_O) \cdot(\mathbf{t}_{I_n}-\mathbf{t}_O) }{\left \| \mathbf{t}_{I_m}-\mathbf{t}_O \right \|  \left \| \mathbf{t}_{I_n}-\mathbf{t}_O \right \|} \right )  
\end{equation}
If $\alpha$ is larger than the preset threshold, then update the training data. As the viewpoints increase, the number of training iterations gradually increases to converge quickly.

\subsubsection{Parallel Training}
We adopt a thread pool approach to enable parallel training of multi-object models. Each worker thread in the thread pool has its own CUDA stream and asynchronously fetches training tasks from the work queue of the object SLAM system. When an object no longer receives new observations, the model will stop training to make more efficient use of computational resources. We configure the pool with 8 worker threads, which is sufficient for most scenes.

\subsection{Volume Rendering}

Differentiable volume rendering is used to optimize the implicit representation of objects. We first transform the camera pose to the object coordinate system and back-project the pixels that are within the object's detection box. If the ray intersects with the 3D bounding box, we compute the truncation distance and sample $N$ points within it. Unlike other RGB-only implicit reconstruction methods, we only perform uniform sampling and do not include the popular importance sampling. This can save the time cost of one inference process of the model, although it slightly compromises the reconstruction quality.

Since we are more interested in dense reconstruction than novel view synthesis, only the positions of the sampled points are encoded and fed into the network to estimate their density values $\sigma $ and colors $c$, without including the ray direction.
For a point $\mathbf{x} _i$, there are its occupancy probability $o_i = 1-exp(-  \sigma_i\delta_i)$ and the probability $w_i = o_i  {\textstyle \prod_{j=1}^{i-1}} (1-o_j)$ that the ray terminates at this point, where $\delta_i = d_{i+1}-d_i$ is the difference in distance between adjacent sample points.
Finally, the predicted color and depth of the corresponding ray $\mathbf{r}$ are defined as follows:
\begin{equation}
\hat{C} (\mathbf{r} )=\sum_{i=1}^{N} w_ic_i,    \qquad   \hat{D} (\mathbf{r} )=\sum_{i=1}^{N} w_id_i
\end{equation}

Object reconstruction requires suppressing the background and occluders since objects are not isolated but embedded in scenes. Instance masks contain valid spatial semantic information that is used to guide learning the geometry distribution of objects and their surroundings. We follow the strategy in \cite{abou2022implicit} and categorize the optimization loss of rays. Specifically, we classify the sampled rays into three types according to their instance masks.
For rays $\mathcal{R}_o$ that hit the reconstructed object, i.e., their mask value $m$ matches with object instance $M_o$, we calculate their photometric loss as usual:
\begin{equation}
L_{rgb} = \sum_{i\in \mathcal{R}_o } {\left \| \hat{C}(\mathbf{r}_i ) - C(\mathbf{r}_i )  \right \| }_2
\end{equation}
For some rays $\mathcal{R}_d$ that have depth supervision, additional depth loss is also included:
\begin{equation}
L_{depth} = \sum_{i\in \mathcal{R}_d } \left | \hat{D}(\mathbf{r}_i ) - D(\mathbf{r}_i ) \right | 
\end{equation}

Second, we expect the space outside the objects to be empty, i.e., the rays pointing to them should not terminate. To achieve this, we give those rays $\mathcal{R}_b$ corresponding to the background varying random colors as supervision to guide them to learn zero density:
\begin{equation}
L_{rr} = \sum_{i\in \mathcal{R}_b } {\left \| \hat{C}(\mathbf{r}_i ) - C_{random}  \right \| }_2
\end{equation}
However, the convergence of volume density caused by this loss is slow and requires many rounds of optimization. We propose an efficient and aggressive loss that skips volume rendering and directly optimizes the density of sampling points as follows:
\begin{equation}
L_{density} = \sum_{i\in \mathcal{R}_b } \sum_{j=1 }^{N} \left |\sigma_j^{\mathbf{r}_i} - 0  \right | 
\end{equation}
Although it looks inelegant, it quickly improves the convergence speed and helps reduce depth ambiguity caused by only monocular images. Fig. \ref{fig3} demonstrates an example where rays pointing to the object in image $I_k$ can rapidly focus on optimizing near the object due to zero density learned by background rays in image $I_j$. 
Finally, for rays that hit other occluding objects, we do not construct optimization loss because we cannot specify spatial information along their paths.
Overall, the total loss for object instance is defined as:
\begin{equation}
L = L_{rgb} + L_{rr} + \lambda_1 L_{depth} + \lambda_2 L_{density}
\end{equation}
where $\lambda_1$ and $\lambda_2$ are loss weights.

\section{EXPERIMENTS}

We evaluate the proposed pipeline on synthetic and real-world datasets. Due to the low requirement of object localization for NeRF training, i.e., the estimated bounding box only needs to loosely enclose objects, we focus on evaluating shape reconstruction. We also provide detailed runtime analysis and two ablation studies that supports our design choices. Considering the online nature of our method, please see the attached video demonstration.

\renewcommand{\arraystretch}{1.1} 
\begin{table}[t]
	\centering
	\caption{Quantitative evaluation of object reconstruction. \textbf{Bold} and \underline{underline} indicate the best and the second-best respectively.}
	\label{table1}
	\resizebox{\columnwidth}{!}{%
		\begin{tabular}{@{}ccccc@{}}
			\toprule
			& Acc. {[}cm{]$\downarrow$} & Comp. {[}cm{]$\downarrow$} & \begin{tabular}[c]{@{}c@{}}Comp.Ratio\\ {[}\textless{}0.4cm \%{]$\uparrow$}\end{tabular} & \begin{tabular}[c]{@{}c@{}}Comp.Ratio\\ {[}\textless{}1cm \%{]$\uparrow$}\end{tabular} \\ \midrule
			{[}33{]}* w/ GT depth & \textbf{0.259} & \textbf{0.162} & \textbf{93.69} & \textbf{99.98} \\
			COLMAP & 0.612 & 0.656 & 57.72 & 81.17 \\
			{[}33{]}* & 0.476 & {\ul 0.228} & {\ul 83.16} & {\ul 99.38} \\
			\textbf{Ours} & {\ul 0.431} & 0.248 & 80.85 & 98.93 \\ \bottomrule
		\end{tabular}%
	}
\end{table}

\begin{figure}[t]
	\centering
	\includegraphics[width=0.93\linewidth]{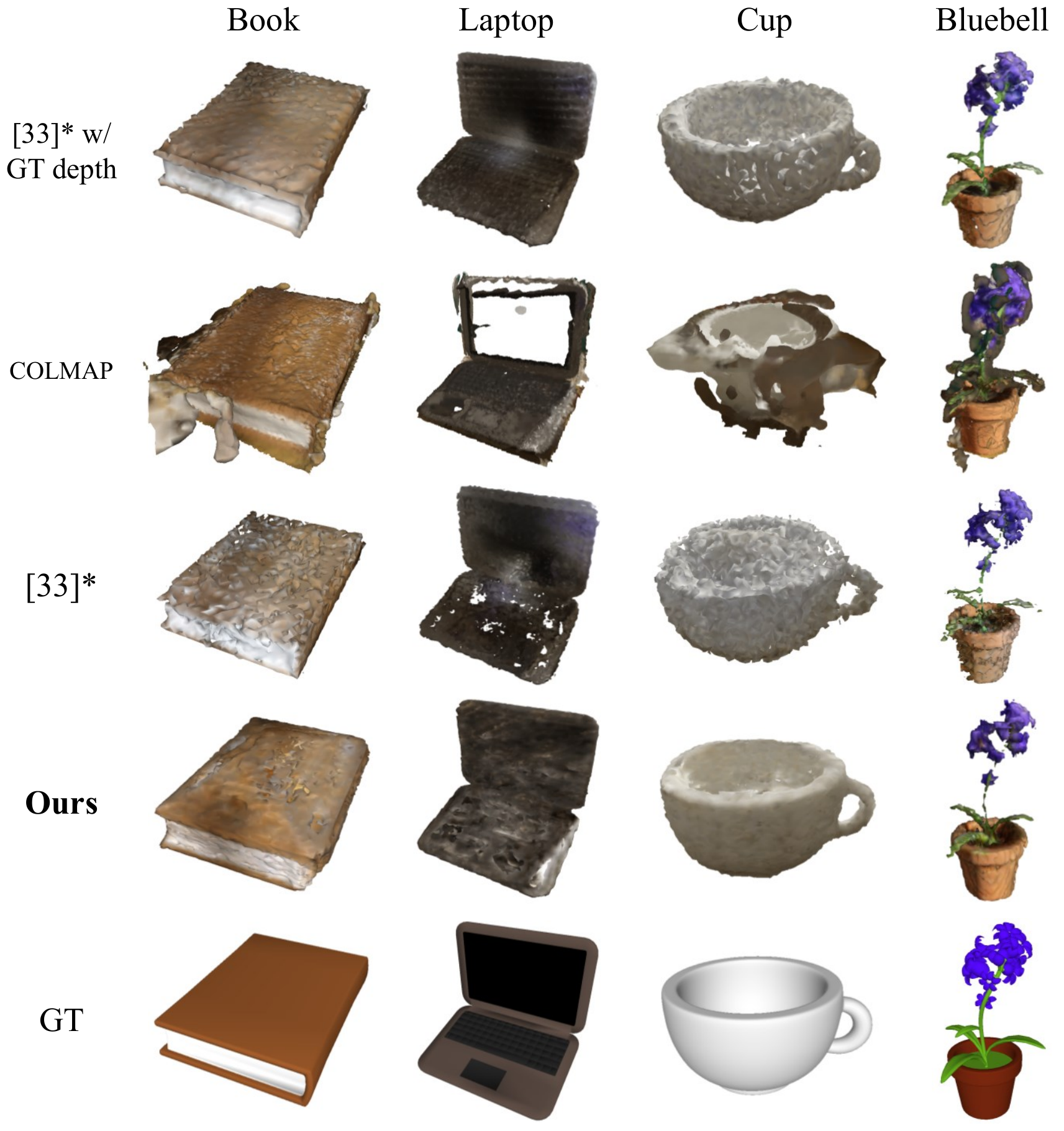}
	\caption{Qualitative results on the Cube-Diorama \emph{room} sequence. Note that all the comparison methods run offline.}
	\label{fig4}
\end{figure}

\subsubsection{Implementation Details}
Our pipeline is implemented using C++ and CUDA, and all experiments are conducted on a desktop computer with a 3.0GHz Intel Xeon 6154 CPU and an NVIDIA RTX 4090 GPU.
All object instances use the same NeRF model parameters, including hash table size $T=2^{16}$, finest resolution $N_{max}=2048$, and hidden size 64 of the single-layer MLP. For volume rendering, we trigger 300 iterations of training each time new observations are received. Each iteration randomly samples 4096 rays from all training images, with $N=32$ sampling points per ray. In addition, we set the loss weights $\lambda_1=0.5$, $\lambda_2=0.01$ and the training data update threshold $\alpha=25^{\circ}$. Marching Cubes \cite{lorensen1987marching} are used to extract meshes online, with the same resolution of $64^3$ for all objects.
\subsubsection{Baselines}
We compare to the classical COLMAP \cite{schoenberger2016sfm} and an implicit object reconstruction method \cite{abou2022implicit} also based on NeRF. We faithfully re-implement the latter, denoted as \cite{abou2022implicit}*. Since both of these comparison methods operate offline and do not consider object localization, we use the training data obtained from running our method online as their input to ensure a fair comparison. Additionally, we also compare our method with two other NeRF-based approaches, namely iMAP \cite{sucar2021imap} and vMAP \cite{kong2023vmap}.

\begin{table}[t]
	\centering
	\caption{Quantitative evaluation of object reconstruction.}
	\renewcommand{\arraystretch}{1.2} 
	\label{table2}
	\resizebox{\columnwidth}{!}{%
		\begin{tabular}{@{}cccccc@{}}
			\toprule
			&  & Acc. {[}cm{]$\downarrow$} & Comp. {[}cm{]$\downarrow$} & \begin{tabular}[c]{@{}c@{}}Comp.Ratio\\ {[}\textless{}1cm \%{]$\uparrow$}\end{tabular} & \begin{tabular}[c]{@{}c@{}}Comp.Ratio\\ {[}\textless{}5cm \%{]$\uparrow$}\end{tabular} \\ \midrule
			\multirow{2}{*}{iMAP} & room-0 & 3.02 & 1.71 & 52.57 & 93.72 \\
			& office-1 & 2.62 & 2.58 & 48.93 & 91.09 \\ \midrule
			\multirow{2}{*}{vMAP} & room-0 & \textbf{2.18} & 1.13 & \textbf{74.09} & 96.68 \\
			& office-1 & \textbf{2.27} & 1.77 & 65.24 & 92.94 \\ \midrule
			\multirow{2}{*}{\textbf{Ours}} & room-0 & 3.65 & \textbf{0.93} & 69.25 & \textbf{98.53} \\
			& office-1 & 3.74 & \textbf{1.15} & \textbf{67.93} & \textbf{97.73} \\ \bottomrule
		\end{tabular}%
	}
\end{table}

\begin{figure}[t]
	\centering
	\includegraphics[width=0.94\linewidth]{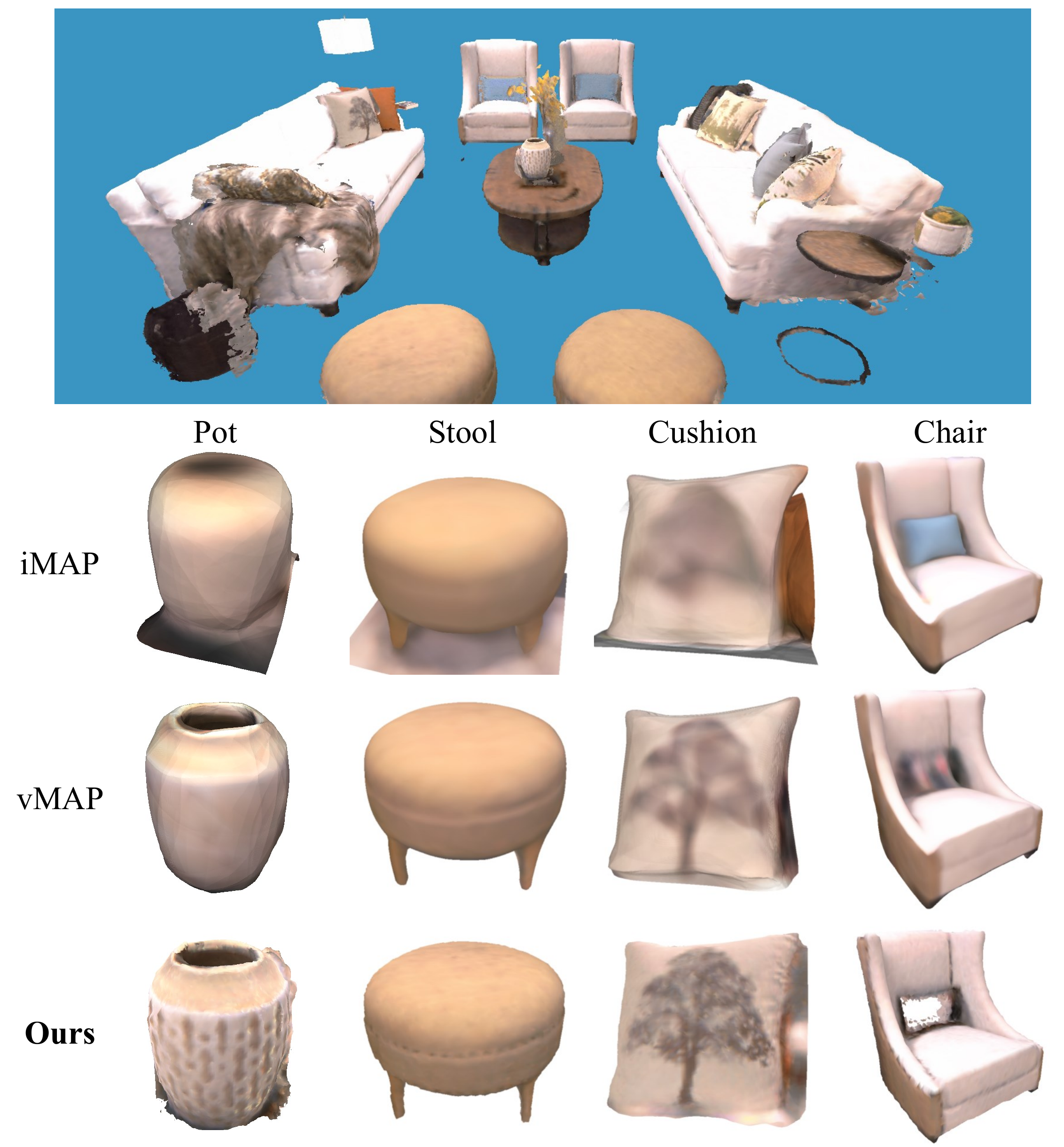}
	\caption{Qualitative results on the Replica \emph{room-0} sequence.}
	\label{fig5}
\end{figure}

\subsubsection{Datasets and Metrics}

We first evaluate on the synthetic Cube-Diorama dataset \cite{abou2022implicit} and Replica dataset \cite{straub2019replica}, which provide ground-truth depths and instance segmentations. \emph{Accuracy}, \emph{Completion} and \emph{Completion Ratio} are used for quantitative evaluation of object reconstruction. On the Cube-Diorama dataset, due to the scale ambiguity of monocular SLAM systems, we use ICP \cite{girardeau2016cloudcompare} to align the reconstructed meshes and the GT meshes. Subsequently, we qualitatively evaluate on a collected real-world dataset.

\subsection{Object Reconstruction Evaluation}

\subsubsection{Cube-Diorama}

We first evaluate the quality of object reconstruction on the \emph{room} sequence, as shown on the left of Fig. \ref{fig1}, which captures four different-shaped objects on a desktop. 
We use ground-truth instance masks to test the upper bound of system performance and additionally show the results of \cite{abou2022implicit} trained with ground-truth depth for reference. Table \ref{table1} presents the quantitative results. Benefiting from the powerful capabilities of implicit representation and volume rendering, our method significantly outperforms traditional COLMAP, which is difficult to handle black or textureless object surfaces. Since both our method and \cite{abou2022implicit} utilize the same network models from \cite{tiny-cuda-nn}, they demonstrate comparable performance. 
However, \cite{abou2022implicit} solely implements offline reconstruction, which requires estimating the poses of all frames and manually labeling object bounding boxes in advance. In contrast, our complete pipeline, integrating object SLAM for online operation, object pose estimation, and parallel training, achieves higher completeness and practicality.
As expected, the use of additional ground-truth depth in \cite{abou2022implicit} makes it easier to capture geometric information, resulting in better results. Fig. \ref{fig4} shows the visualization results of all objects. Our method can generate watertight object meshes. It is worth noting that, due to the reflection and the noise in the estimated camera pose, all RGB-only methods suffer from artifacts. This remains a challenge for monocular reconstruction without 3D priors.

%Nonetheless, our complete pipeline, integrating object SLAM for online operation, object pose estimation, and parallel training, achieves higher completeness and practicality. 

\subsubsection{Replica}

Since our heuristic object pose estimation method based on monocular input is difficult to handle large objects such as beds and dining tables, we restrict the evaluation of RO-MAP to the \emph{room-0} and \emph{office-1} sequences that mainly contain smaller objects. The number of objects in these sequences is 13 and 29, respectively. For dense reconstruction, we use RGB-D input similar to the comparison methods. The quantitative results are presented in Table \ref{table2}. Our method achieves higher reconstruction completion. Due to our straightforward uniform sampling strategy, the sampling number near the surface of larger objects is insufficient, which leads to some artifacts inside the generated object surfaces and consequently impacting the accuracy metrics of our method. Fig \ref{fig5} illustrates the visualization results, RO-MAP has better reconstruction fineness and can construct multi-object maps with semantic information. Compared to vMAP, our method encounters challenges in dealing with heavily occluded areas, which will be a focus of future research.

\begin{figure}[t]
	\centering
	\includegraphics[width=0.93\linewidth]{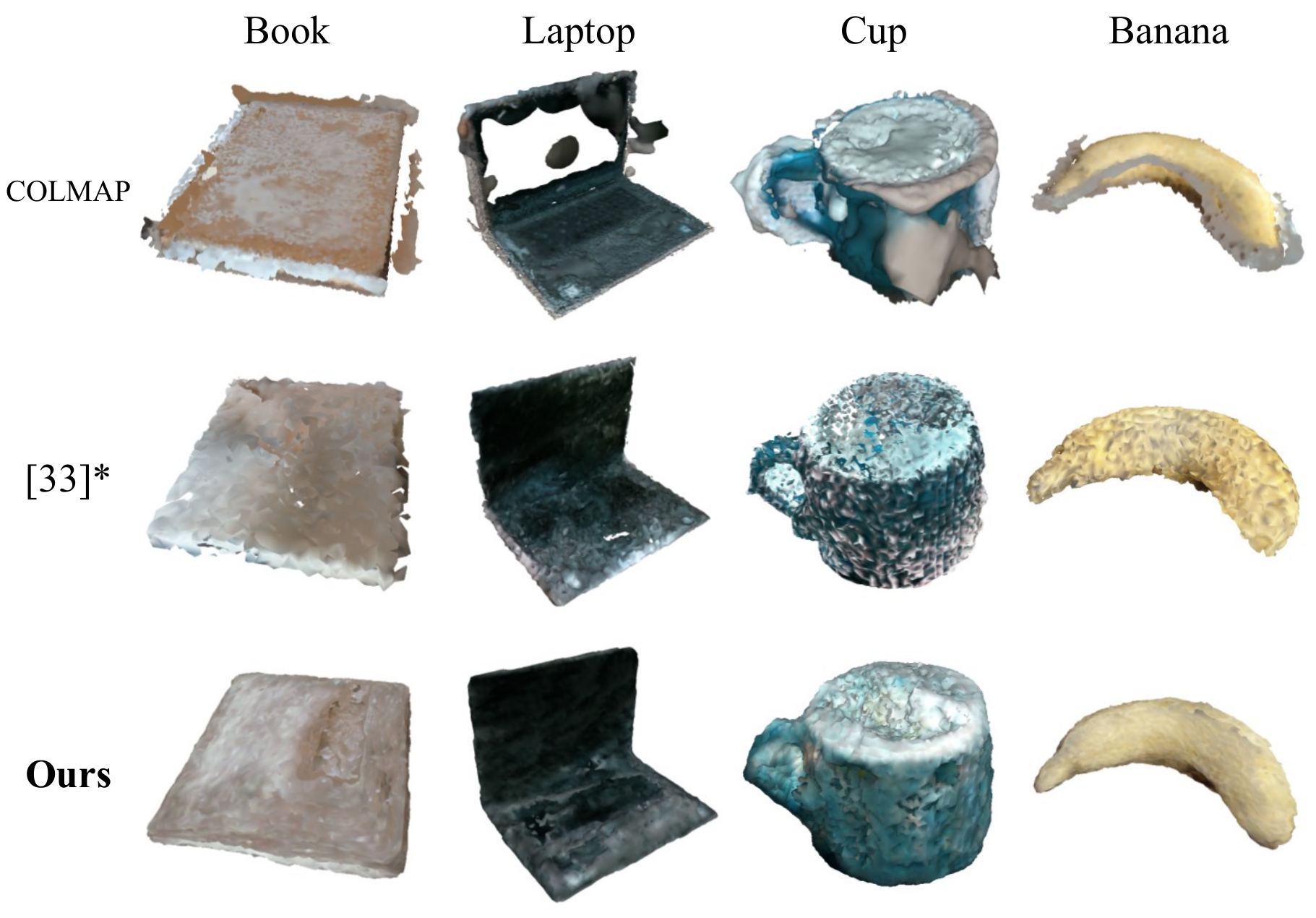}
	\caption{Reconstruction results of all objects on the real-world sequence. Instance masks are produced by YOLOv8.}
	\label{fig6}
\end{figure} 

\subsubsection{Real-world Sequence}
We evaluate on a real scene collected with a Realsense D455 camera, as shown on the right of Fig. \ref{fig1}. Noisy object masks are provided by YOLOv8, which has real-time performance. Fig. \ref{fig6} presents the qualitative results. We can see that COLMAP fails to reconstruct the non-Lambertian laptop screen, resulting in large holes. Compared with \cite{abou2022implicit}, our method generates more complete object reconstruction and has better visual quality.
\subsubsection{Challenging Real-world Sequence}
\begin{figure}[t]
	\centering
	\includegraphics[width=0.97\linewidth]{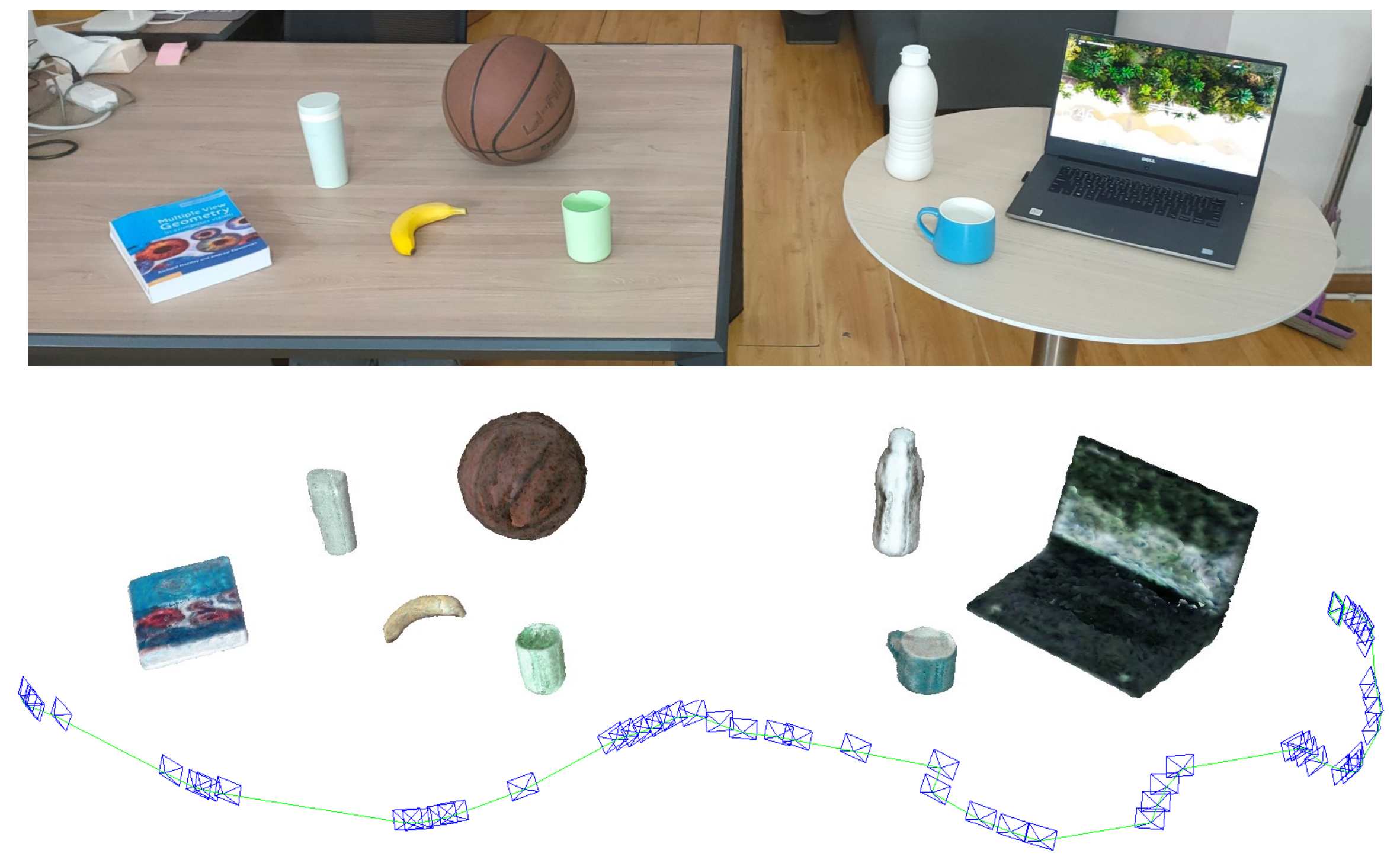}
	\caption{(Top) The scene of the challenging real-world sequence. (Bottom) Dense object map and constrained camera motion trajectory.}
	\label{fig7}
\end{figure}

In practical applications of robotics or AR, it is often impossible to obtain observations from all viewpoints of objects. We provide a challenging real-world sequence, which contains eight objects of different shapes, and the camera only gives limited viewpoint observations along a constrained motion trajectory. Fig. \ref{fig7} shows the scene and the qualitative results. 
It can be observed that the generated object meshes are well separated from the background in the observed viewpoints.
However, for the unobserved regions of the objects, although the interpolation-based multi-resolution feature grid has some predictive ability, it still cannot produce satisfactory results. Reconstructing objects from partial observations \cite{lee2023just} is an interesting direction for future work.

\renewcommand{\arraystretch}{1.35} 

\begin{table}[]
	\centering
	\caption{Runtime of different system components.}
	\label{table3}
	\resizebox{\columnwidth}{!}{%
		\begin{tabular}{|c|cc|cc|}
			\hline
			Component                    & \multicolumn{2}{c|}{Tasks}                                                                                                            & \multicolumn{2}{c|}{Runtime(mSec)}                   \\ \hline
			\multirow{2}{*}{Object SLAM} & \multicolumn{2}{c|}{Frontend Tracking}                                                                                                         & \multicolumn{2}{c|}{41.6}                            \\ \cline{2-5} 
			& \multicolumn{2}{c|}{Backend optimization}                                                                                                          & \multicolumn{2}{c|}{204}                             \\ \hline
			\multirow{4}{*}{NeRF Model}  & \multicolumn{1}{c|}{\multirow{3}{*}{\begin{tabular}[c]{@{}c@{}}Iteration\\ (avg/object)\end{tabular}}} & Ray Sampling                 & \multicolumn{1}{c|}{0.0932} & \multirow{3}{*}{0.706} \\ \cline{3-4}
			& \multicolumn{1}{c|}{}                                                                                  & Forward and Volume Rendering & \multicolumn{1}{c|}{0.156}  &                        \\ \cline{3-4}
			& \multicolumn{1}{c|}{}                                                                                  & Backward and Optimization    & \multicolumn{1}{c|}{0.455}  &                        \\ \cline{2-5} 
			& \multicolumn{2}{c|}{Marching cubes}                                                                                                   & \multicolumn{2}{c|}{2.84}                            \\ \hline
		\end{tabular}%
	}
\end{table}

\subsection{Runtime Analysis}

Table \ref{table3} shows the detailed breakdown of average computation time for each main component. For object SLAM, our hand-crafted object pose estimation method only introduces a small amount of time consumption to the original ORB-SLAM2.
For a single NeRF model, our parallel implementation requires only 0.7ms for one iteration of training.
The number of iterations for training different object instances depends on the size of the observed view angle. On average, each object takes about 2 seconds.
Compared to the whole scene reconstruction, representing and optimizing a single object allows us to use tiny networks and simple sampling strategies, which help reduce the branch divergence problem in parallel computing and improve the training speed. 
When the scene contains a large number of objects, resulting in a queuing situation in the NeRF thread pool, our system's unidirectional data flow ensures that it does not block the SLAM tracking process. It only slightly increases the training time per iteration to approximately 0.83ms.

\subsection{Ablation Study}
\subsubsection{Losses}
\begin{figure}[t]
	\centering
	\includegraphics[width=0.9\linewidth]{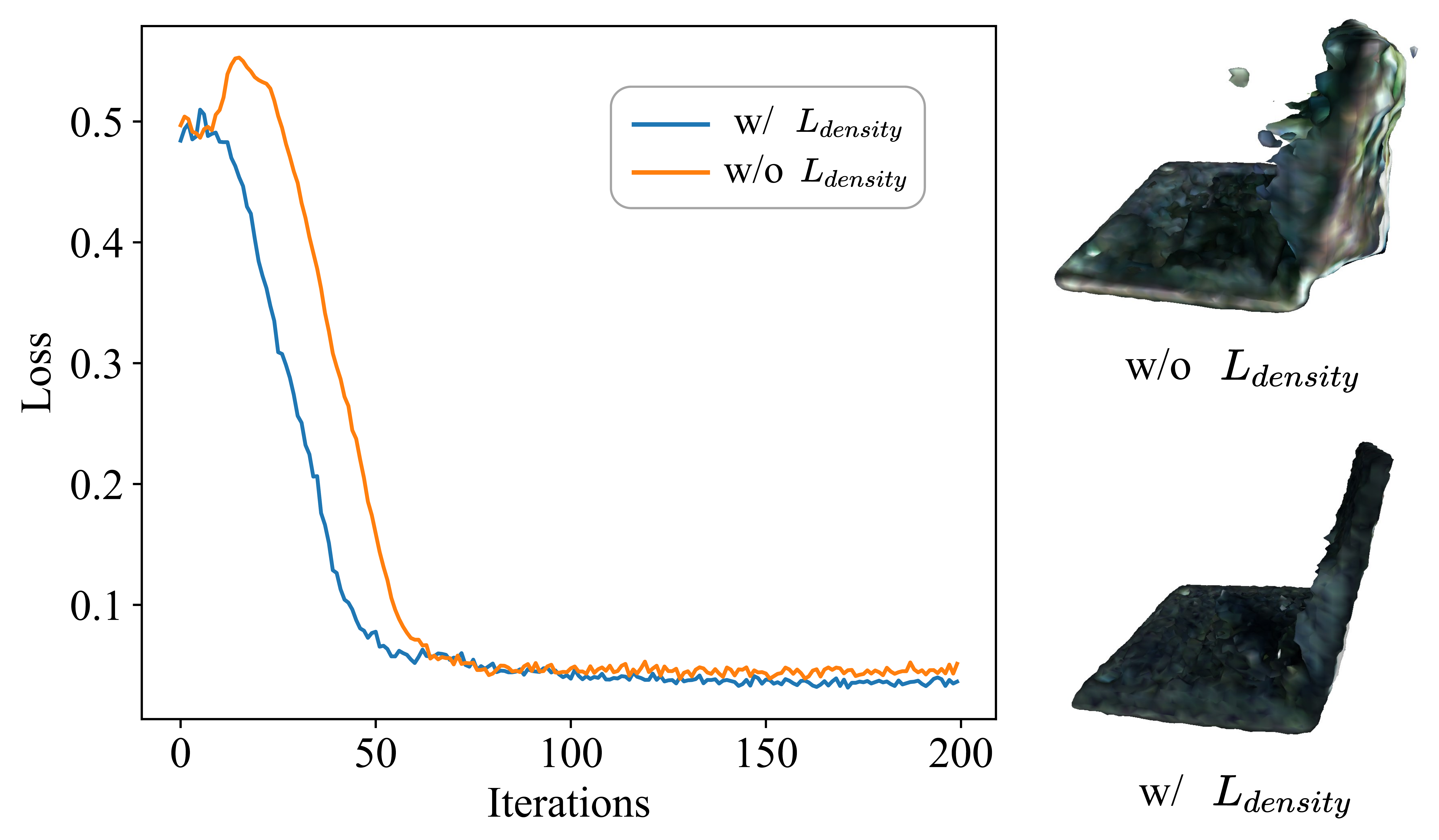}
	\caption{Ablation Study. (Left) Convergence plot of training with all data on the \emph{room} sequence. (Right) Intuitive results of the laptop from the challenging sequence.}
	\label{fig8}
\end{figure}

The depth ambiguity caused by monocular input has a significant impact on learning the geometry of textureless or smooth objects, such as leading to slow convergence and scattered artifacts. The convergence plot in Fig. \ref{fig8} shows the comparison results on the \emph{room} sequence. 
It can be observed that the training guided only by the random color loss is difficult to converge in the early stage, which indicates that directly regressing the voxel density of empty regions is helpful for the disambiguation of the optimization process. The right side of Fig. \ref{fig8} shows an intuitive example.

\subsubsection{Object Model}

We investigate the impact of different object models on the reconstruction quality, including the size of the hash encoding table and the number of layers in MLP.
In Table \ref{table4}, we can see that using larger models did not improve the reconstruction quality. This can be attributed to the limited number of iterations caused by online processing, which is insufficient to adequately train model parameters.
In contrast, models with fewer parameters are already capable of representing individual objects and have faster training speed.

\begin{table}[]
	\centering
	\renewcommand{\arraystretch}{1}
	\caption{Ablation study of object model sizes.}
	\label{table4}
	\begin{tabular}{@{}ccccccc@{}}
		\toprule
		Hash table size & 14 & 16 & 18 & 20 & 16 & 16 \\
		MLP layers & 1 & 1 & 1 & 1 & 2 & 3 \\ \midrule
		Acc. {[}cm{]$\downarrow$} & 0.435 & 0.424 & 0.451 & 0.450 & 0.409 & 0.442 \\
		Comp. {[}cm{]$\downarrow$} & 0.227 & 0.233 & 0.236 & 0.243 & 0.241 & 0.243 \\
		Iteration time {[}ms{]} & 0.658 & 0.706 & 0.921 & 1.495 & 0.862 & 1.052 \\ \bottomrule
	\end{tabular}
\end{table}

\section{CONCLUSIONS}

We present RO-MAP, a real-time multi-object mapping pipeline that only uses monocular input and does not rely on 3D priors. The method employs neural radiance fields as implicit shape representations, and combines them with lightweight object SLAM, to localize and reconstruct objects in a scene and generate dense object map with semantic information. Our high-performance implementation allows creating separate implicit models for each object, which can be incrementally trained and converge quickly. Comprehensive experiments demonstrate the effectiveness and advantages of RO-MAP. In the future, we are interested in how to utilize implicit object maps for downstream tasks such as robot navigation, grasping, and relocalization.

\section{Acknowledgments}

We thank Jad Abou-Chakra for his support on the dataset and Quei-An Chen for helpful discussions.

\bibliographystyle{IEEEtran}
\bibliography{ref}
	
	%\printbibliography
\end{document}